\documentclass{article}

\usepackage{graphicx}
\usepackage{array}
\usepackage{amsmath}
\usepackage{algorithm}
\usepackage{algpseudocode}
\usepackage{tikz}
\usetikzlibrary{arrows.meta,positioning,fit,backgrounds,shadows,calc}
\usepackage{float}
\usepackage{placeins}
\usepackage{microtype}
\usepackage[numbers]{natbib} 
\usepackage[hidelinks]{hyperref} 

\begin{document}

\title{Superlinear Multi-Step Attention}

\author{
    Yufeng Huang\thanks{Corresponding author: yufeng@concavity.ai} \\
    Concavity AI \\
    \texttt{yufeng@concavity.ai}
}
\date{}

\maketitle

\begin{abstract}
In this paper, we propose \textbf{Superlinear attention}, a fully trainable multi-step attention architecture that achieves subquadratic complexity for long sequences
while preserving \textbf{random context access} (a.k.a.\ structural non-exclusion): no eligible token position is structurally excluded from being selected for attention.
Superlinear attention reformulates standard causal self-attention as a multi-step search problem with $N$ steps, yielding an overall complexity of $O(L^{1+\frac{1}{N}})$.
To illustrate the architecture, we present a baseline $N=2$ implementation, which is algorithmically analogous to standard jump search.
In this $O(L^{3/2})$ instantiation, the first step performs $O(L^{3/2})$ span-search to select relevant spans of the sequence, and the second step applies $O(L^{3/2})$ 
span-attention (standard attention restricted to the selected spans). With this configuratoin, we achieve an average decoding throughput of
109 tokens/sec at 1M context length and 76 tokens/sec at 10M context in our implementation on a modified 30B hybrid MoE model on a single B200 GPU. 
With limited training, we also obtain strong performance on the NIAH (Needle In A Haystack) task up to 256K context length,
demonstrating that the routed span selection is learnable end-to-end. 
This paper emphasizes architectural formulation, scaling analysis, and systems feasibility, and presents initial validation; comprehensive quality evaluations across diverse long-context tasks are left to future work.
\end{abstract}

\section{Introduction}
Large language models (LLMs) are increasingly expected to operate 
as \emph{long-context} systems \cite{liu2025comprehensive}: they must read and synthesize many 
documents at once \cite{zou2025docbench}, maintain memory \cite{zhang2025survey} and consistency over long interactive 
or agentic workflows \cite{plaat2025agentic}, reason over large retrieved evidence sets (RAG) \cite{lewis2020retrieval, singh2025agentic}, 
and understand or edit large codebases and logs without losing critical 
details \cite{jiang2024survey}. As context windows extend from thousands to hundreds of thousands 
or even millions of tokens, these applications become feasible in a single pass, 
enabling more reliable retrieval, long-horizon reasoning, and test-time ``scratchpad'' 
style deliberation within one context. However, the quadratic cost $O(L^2)$
of standard causal self-attention \cite{vaswani2017attention} remains a fundamental bottleneck, 
limiting scalability to extremely long sequences.

Ever since the introduction of the transformer architecture, there have been 
numerous efforts to improve the efficiency of self-attention mechanisms. 
To reduce the $O(L^2)$ complexity of self-attention, early approaches have 
focused on sparse attention mechanisms, where the idea 
is to modify the attention mask to attend to only a subset of tokens \cite{child2019generating, 
beltagy2020longformer, zaheer2020big, xiao2023efficient, jiang2024minference}. However, 
these methods often sacrifice random context access
by imposing hard restrictions on which token positions can be selected for attention, 
which can be detrimental to model performance.

Recent works have explored alternative approaches to improve efficiency while maintaining 
random context access \cite{hu2024efficient, leng2025understanding}.
One notable example is the block-based retrieval approach, where the model retrieves relevant blocks 
of tokens from a large context window \cite{mohtashami2023landmark}. 
This can be aligned with hardware level blocks to achieve efficient implementation 
\cite{yuan2025native, yan2025fsa, lu2025moba, xiao2025optimizing}.
While this method maintains random context access, and the final attention computation is linear in complexity,  
the retrieval step can still be $O(L^2)$ in the worst case for exact scoring, since it may compare 
the query against all tokens (or all blocks) in the sequence to identify relevant blocks. Similar scaling issues
also arise in token-level retrieval approaches like top-$k$ sparse attention \cite{liu2025deepseek, 
xiu2025preliminary}, where the overall complexity can remain $O(L^2)$ when exact token/block scores must be computed to identify relevant tokens.
Approximate indexing or hierarchical retrieval can reduce this cost in many settings \cite{kitaev2020reformer, roy2021efficient, zhu2021h, zeng2022multi, wu2022memorizing}, but introduces additional assumptions/approximation and is a separate design axis.
Empirically, such retrieval-style approaches can be computationally competitive with optimized dense attention kernels (e.g., FlashAttention) \cite{dao2022flashattention, dao2023flashattention}
and can retain strong performance in part due to random context access. However, the quadratic complexity of 
the retrieval step limits its scalability for extremely long sequences like 100M tokens or even 1B tokens. 

The main challenge remains to achieve subquadratic complexity while maintaining random context access.
We propose \textbf{Superlinear attention}, a multi-step attention architecture that addresses this challenge.
The mechanism reformulates attention as a multi-step search problem, 
allowing us to leverage efficient search algorithms to improve performance. In conventional search algorithms
like jump search, we can achieve $O(\sqrt{L})$ search complexity by searching over 
$O(\sqrt{L})$ anchors and then performing linear search within the selected block.

Prior work explored an $O(L^{1+p})$ attention mechanism that attends to $O(L^p)$ tokens 
by building on top of the linear recurrence layers in a hybrid architecture. Specifically, we modified the full attention 
layer in the hybrid architecture \cite{blakeman2025nemotron} to attend to only $O(L^p)$ tokens,
where long-range dependencies are captured by the underlying linear recurrence (Mamba-2) layers. 
Although this approach suffers from random context access issues just like the other methods discussed above, it provides a 
foundation for developing a subquadratic search mechanism. Instead of attending to only $O(L^p)$ tokens directly,
we can use the $O(L^p)$ attention to find relevant spans of $O(L^{1-p})$ tokens,
so that we retain \textbf{random context access} while achieving subquadratic complexity. This process can be generalized to $N$ steps,
resulting in an overall complexity of $O(L \cdot L^{1/N})$.

In this study, we introduce the Superlinear attention architecture, which consists of four main components:
accumulation, span-search, span-attention, and combination. We provide a detailed description of each component and how they 
work together to achieve subquadratic complexity while maintaining random context access. In order to validate the architecture, 
we present a baseline $N=2$ implementation, which achieves $O(L^{3/2})$ complexity. 
We also provide experimental results demonstrating the effectiveness of the proposed architecture in terms of
scaling and learnability. The main focus of this paper is to introduce the architecture and provide initial validation.
Optimization of the implementation and extensive empirical evaluation are left for future work.

\vspace{1em}

\noindent\textbf{Important clarification:} ``Random context access'' does \emph{not} mean attending to all tokens for each query. Rather, it means any eligible token position \emph{can} be selected for attention by the content-dependent span-search/routing mechanism (i.e., no hard-coded sparsity pattern structurally excludes it).

\vspace{0.5em}
\noindent\textbf{Scope of claims.} This paper makes two primary claims: (i) \emph{architectural/theoretical}: a multi-step span-search + span-attention mechanism can achieve subquadratic attention complexity while preserving random context access (structural non-exclusion), together with scaling analysis and conditions for practical instantiations; and (ii) \emph{systems}: with appropriate kernel design, the resulting irregular span pattern can be implemented efficiently enough to yield practical long-context speedups. We present initial evidence for both, while leaving broad quality benchmarking and extensive ablations to future work.

\vspace{0.5em}
\noindent\textbf{Contributions.} We summarize the main contributions of this work as follows:
\begin{itemize}
    \item \textbf{Random-context-access-preserving formulation.} We formalize \textbf{random context access} (structural non-exclusion) as a design criterion for long-context attention and propose a multi-step span routing mechanism that preserves it.
    \item \textbf{Scaling analysis.} We analyze the compute scaling of the multi-step mechanism and derive the $N=2$ balanced setting with $O(L^{3/2})$ span-search and span-attention.
    \item \textbf{Practical instantiation.} We describe a baseline $N=2$ design (anchor schedule, span construction, and differentiable top-$k$ combination) that can be integrated into existing hybrid transformer architectures.
    \item \textbf{Kernel design and feasibility evidence.} We present a bucketed GPU kernel strategy for irregular spans and report initial long-context throughput and learnability evidence.
\end{itemize}

\section{Methodology}

\subsection{Architecture overview}

Our goal is to design an architecture that can efficiently 
process long sequences while maintaining random context access.
By developing a multi-step attention mechanism, we can achieve subquadratic complexity 
while still allowing attention to be directed to \emph{any} eligible token position. 
Intuitively, this is analogous to search procedures that narrow down candidates in stages, except the selection is content-dependent and learned.

\begin{figure}[H]
    \centering
    \makebox[\textwidth][c]{%
    \includegraphics[width=1.1\textwidth]{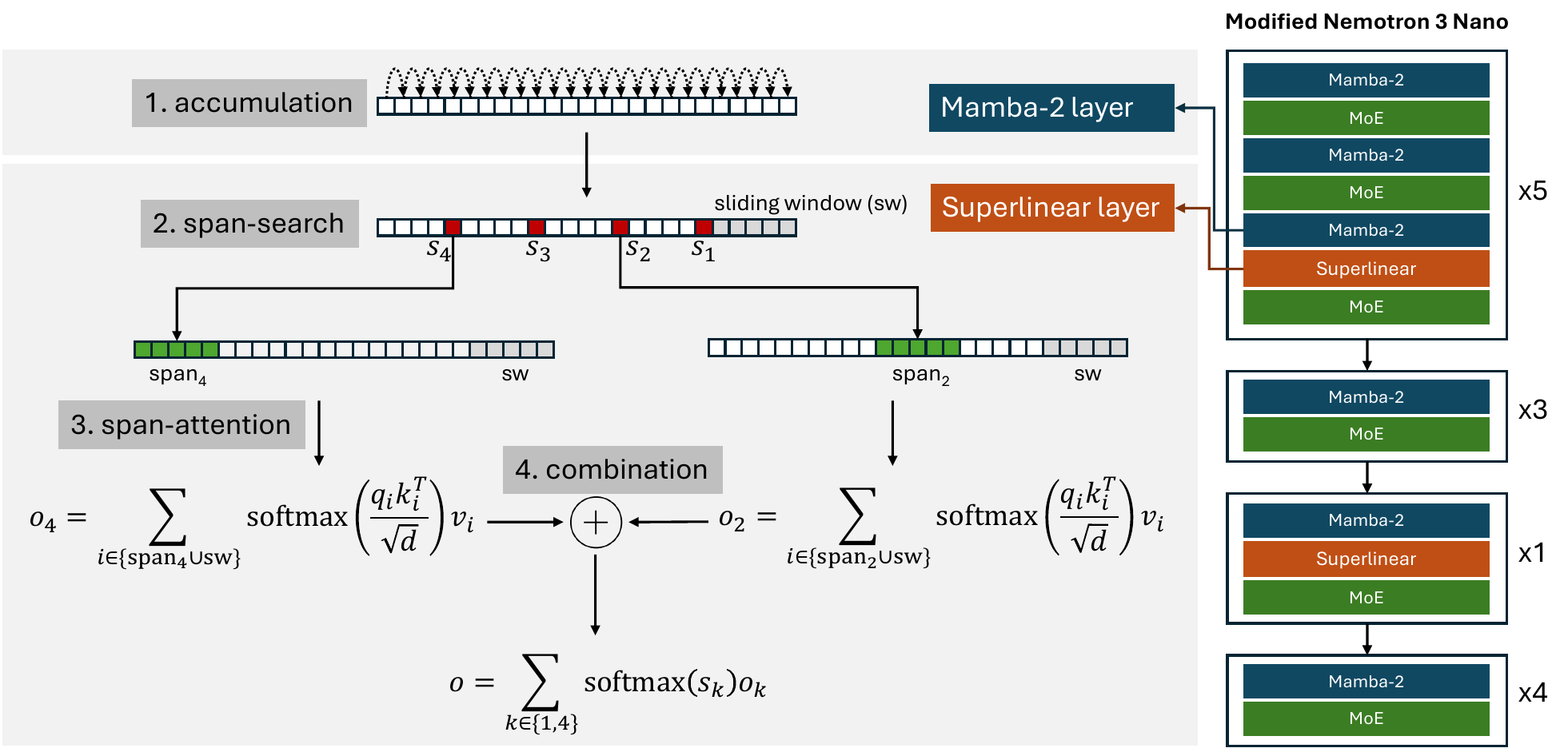}
    }
    \caption{Left: Overview of the proposed Superlinear attention architecture. 
    The architecture consists of four main components: accumulation, span-search, span-attention, and combination. 
    Each component plays a crucial role in achieving subquadratic complexity while maintaining \textbf{random context access}.
    In the example, we search over 4 spans, and top-$2$ spans are selected for attention and combined for the final output.
    The schematic in this figure only shows the 1D attention mask for a single query token for simplicity; 
    Figure~\ref{fig:span_search_and_attention} shows the full 2D attention map
    for the span-search and span-attention steps.
    Right: overall model architecture with the Superlinear attention layer integrated into a hybrid transformer architecture 
    using Nemotron-3-Nano \cite{nvidia2025nemotron3nanoopen, nvidia2025nvidianemotron3efficient} as the base model. All the standard attention layers in the base model are
     replaced by Superlinear attention layers.
    }
    \label{fig:superlinear_overview}
\end{figure}

Throughout this study, we assume a standard causal eligibility constraint ($j\le i$) 
together with any task-specific masking (e.g., padding). Random context access means that, 
under the mechanism, no eligible position is structurally excluded by a fixed sparsity pattern.

Our architecture separates (i) building a per-position \emph{representative} stream that summarizes the prefix and 
(ii) using that stream to route attention to a small number of contiguous spans. 
The representative stream can be produced by an accumulation component 
such as linear attention \cite{katharopoulos2020transformers} or SSM \cite{gu2021efficiently}---which are unified under the same linear recurrence framework \cite{dao2024transformers, team2025kimi}---and can be cached.

In addition to subquadratic complexity and random context access, the architecture must also be fully trainable via backpropagation. 
Otherwise, the model cannot learn to use the attention mechanism. 
To achieve this, we leverage a differentiable combination mechanism inspired by Mixture-of-Experts (MoE) routing \cite{shazeer2017outrageously, fedus2022switch}, described in Section~\ref{sec:combination}.

Finally, computational efficiency is a key consideration in the design of the architecture. Many subquadratic attention mechanisms have been proposed in recent years,
but because of the efficiency of FlashAttention \cite{dao2022flashattention, dao2023flashattention, ye2025flashinfer} and the efficient usage of hardware accelerators, 
the efficiency advantage of these subquadratic methods has diminished. 
Therefore, the proposed architecture must also admit efficient implementations on modern hardware. 
For the search step over sparse tokens, rather than 
using tokens at arbitrary locations, we can use a fixed stride pattern to access the tokens.
 The fixed stride pattern allows us to calculate the search scores 
in a more efficient manner. For the attention step over contiguous spans, 
the memory access pattern is already efficient since the spans are contiguous in memory.
Because the attention mechanism is able to attend to any eligible token position, 
the full KV cache must be retained during inference. 
For our experimentation up to 10M context length, the KV cache memory usage has 
not been a bottleneck, since the base model \cite{nvidia2025nemotron3nanoopen, nvidia2025nvidianemotron3efficient} we are using 
only requires approximately 6GB memory for each million tokens.

Based on these considerations, we propose the Superlinear attention architecture, which consists of four main components: 
accumulation, span-search, span-attention, 
and combination. The accumulation component leverages linear recurrence mechanisms (e.g., linear attention or SSM) to efficiently
process long sequences and accumulate relevant information up to each token. The span-search component 
identifies top-$k$ spans of the sequence based on the accumulated information. This step can be repeated $N-1$
times, each time reducing the number of tokens to attend to by a factor of $L^{1/N}$. 
Next, the span-attention component performs standard attention on the top-$k$ identified spans of length $L^{1/N}$, allowing 
the model to focus on the most relevant parts of the sequence.
Finally, the combination component combines the top-$k$ attended spans using softmax-weighted gating, 
which makes the span-search trainable via backpropagation.

In the baseline $N=2$ implementation considered in this feasibility study, the following choices are fixed hyperparameters: the anchor pattern $\textsc{Anchors}(i)$ (a deterministic function of $i$), the number of routed spans $k$, the span extents $(b,f)$ (which set the span length, as defined below), and the optional local window size $w$. The learned parts are standard projection matrices for $Q,K,V$ and the routing/query projection that produces $Q_s$.

The roles of the components are summarized as follows:
\begin{itemize}
    \item \textbf{Accumulation}: Efficiently process long sequences and accumulate relevant information. $O(L)$ complexity for this step using linear recurrence (linear attention/SSM). 
    \item \textbf{Span search}: Identify top-$k$ spans of the sequence. $O(L^{1+1/N})$ complexity for this step using $N-1$ span-search steps.
    \item \textbf{Span attention}: Perform standard attention on the identified spans. $O(L^{1+1/N})$ complexity for this step.
    \item \textbf{Combination}: Combine the attended spans in a trainable manner. $O(L)$ complexity for $k$ spans for each token.
\end{itemize}

\subsection{Comparison with existing architectures}

We compare our proposed architecture with existing architectures in terms of search complexity, 
attention complexity, random context access, and total time complexity. As shown in Table~\ref{tab:architecture_comparison},
the conventional approach performs single-step calculation, and random context access depends directly on the attention complexity.

In recent years, various multi-step approaches have been proposed to improve efficiency \cite{yuan2025native, yan2025fsa, lu2025moba, xiao2025optimizing, liu2025deepseek, xiu2025preliminary}. While these methods are 
able to achieve random context access, and calculate attention efficiently by reducing the overall overhead, the 
search complexity remains quadratic in sequence length. 

Another notable approach is clustering-based attention, which is also a multi-step method 
that reduces the search complexity to sublinear levels \cite{kitaev2020reformer, roy2021efficient, tay2020sinkhorn}. 
However, this method still faces challenges in achieving random context access, as it only 
attends to a subset of tokens in the sequence due to the clustering process. In these methods, tokens are assigned to fixed clusters and attention is restricted within each cluster.
This structurally blocks cross-cluster access, so random context access is not satisfied even if clusters are large and the within-cluster reachable set is large (up to $O(L)$).

Our proposed Superlinear attention architecture addresses these limitations by achieving subquadratic complexity
while maintaining random context access. 
Future work can explore higher values of $N$ to further reduce complexity.
Increasing $N$ reduces the exponent in $O(L^{1+1/N})$ but also increases the number of span-search steps.
A separate direction is to use a binary or $k$-ary search strategy with $O(\log L)$ steps, which yields $O(L \log L)$ complexity.

\begin{table}[H]
\centering
\small
\makebox[\textwidth][c]{%
\resizebox{1.3\textwidth}{!}{%
{\setlength{\extrarowheight}{3pt}\renewcommand{\arraystretch}{1.2}%
\begin{tabular}{|l|l|l|l|l|}
\hline
\shortstack[t]{Architecture} & \shortstack[t]{Search/Clustering} & \shortstack[t]{Attention} & \shortstack[t]{Random context access\\(explicit reachability)} & \shortstack[t]{Overall complexity} \\
\hline
RNN/Linear Attention/SSM & - & $O(L)$ & Partial (reachable $O(1)$) & $O(L)$ \\
PPA (our previous work) & - & $O(L^{1+p})$ & Partial (reachable $\sim O(L^{p})$)  & $O(L^{1+p})$ \\
Transformer & -  & $O(L^2)$ & \textbf{Yes} (reachable $O(L)$) & $O(L^2)$ \\
Streaming/Sliding Window & - & $O(L)$ & Partial (reachable $O(1)$) & $O(L)$ \\
Retrieval approach & $O(L^2)$ & $O(L)$ & \textbf{Yes} (reachable $O(L)$) & $O(L^2)$ \\
Clustering approach & $O(L^{3/2})$ or $O(L\log L)$ & $O(L)$ & Partial (reachable $O(L)$ within cluster) & $O(L^{3/2})$ or $O(L \log L)$ \\
Superlinear attention (2-step) & $O(L^{3/2})$ & $O(L^{3/2})$ & \textbf{Yes} (reachable $O(L)$) & $O(L^{3/2})$ \\
Superlinear attention (Multi-step) & $O(L^{1+1/N})$ & $O(L^{1+1/N})$ & \textbf{Yes} (reachable $O(L)$) & $O(L^{1+1/N})$ \\
\hline
\end{tabular}}%
}%
}
\caption{Comparison of different attention architectures for training and prefilling. 
Search complexity refers to the computational cost of identifying relevant tokens.
 Attention complexity refers to the cost of computing attention over selected tokens. 
 Random context access (a.k.a.\ structural non-exclusion) indicates whether any eligible key position can, in principle, be selected as an individual key index by the content-dependent mechanism (explicit reachability), i.e., no fixed sparsity pattern permanently rules it out. We use \textbf{Yes} for architectures that satisfy structural non-exclusion, and \textbf{Partial} for architectures that allow a nontrivial but structurally limited explicitly reachable set.
 We additionally use reachable-set size ($O(\cdot)$) as a coarse proxy for how many key positions are explicitly accessible per query under the mechanism.
 Overall complexity is determined by the dominant operation for training/prefilling (i.e., processing $L$ queries). For incremental decoding (one query per step), the relevant quantity is the per-token (single-query) cost: dense attention becomes $O(L)$ per generated token, while in the $N=2$ Superlinear setting routing+attention scales as $O(L^{1/2})$ per generated token (up to constant factors), with other cached recurrent/MLP components contributing $O(1)$ per token.}
\label{tab:architecture_comparison}
\end{table}

\subsection{Random context access (structural non-exclusion)}
We use \emph{random context access} to mean \emph{structural non-exclusion}: for any query position $i$, no eligible key position $j\le i$ is permanently excluded by a fixed sparsity pattern.
The criterion is distinct from attending to all tokens for each query; instead, it requires that every key position remains \emph{reachable} under the (content-dependent) span-search/routing mechanism.
As an architectural property, random context access permits access in principle, although a learned router may still concentrate probability mass on a small subset of spans.
This criterion makes it clear why many efficient attention patterns trade off random context access.
Sliding-window/streaming attention enforces a fixed receptive field, so most positions become unreachable once they fall outside the window.
Clustering- or hashing-based methods often restrict tokens to fixed partitions, which prevents cross-partition access unless additional global mixing is introduced.
Random sampling can improve average reach but does not guarantee that any particular key is reachable for a given query.

Superlinear attention achieves random context access by combining (i) a sublinear set of candidate anchors (search tokens) with (ii) a span-construction rule that assigns a contiguous \emph{candidate span} to each anchor.
In the baseline $N=2$ setting, the stride pattern yields $O(i^{1/2})$ anchors for query $i$, and choosing a span length scale proportional to $i^{1/2}$ ensures that the \emph{family of candidate spans} (one per anchor, plus an optional local window) collectively covers all keys $j\le i$ (Figure~\ref{fig:structural_non_exclusion}).
The router then selects a small subset (top-$k$) of these candidate spans to actually attend to; random context access means no key is structurally unreachable, not that every key is attended for every query.
Both the number of anchors and the span length are sublinear, so search and attention can remain subquadratic while avoiding hard-coded exclusion.

\subsection{Two-step mechanism and scaling analysis}

For the 2-step mechanism, we parameterize span-search and span-attention with tunable exponents $p_{sh}\in(0,1)$ and $p_{sp}\in(0,1)$.
In span-search, each query scores a sublinear set of $O(L^{p_{sh}})$ anchor tokens, costing $O(L\cdot L^{p_{sh}})=O(L^{1+p_{sh}})$ over $L$ queries.
In span-attention, each query attends to routed contiguous spans with total size $O(L^{p_{sp}})$, costing $O(L\cdot L^{p_{sp}})=O(L^{1+p_{sp}})$.
Therefore, the overall training/prefill complexity of the 2-step mechanism is
$\max\{O(L^{1+p_{sh}}),\, O(L^{1+p_{sp}})\}$.

To ensure random context access (structural non-exclusion), the family of candidate spans must be able to cover all eligible keys. 
In the 2-step setting, this requires that the number of anchors per query ($\sim L^{p_{sh}}$)
 times the span length scale ($\sim L^{p_{sp}}$) is at least linear in $L$, 
 which yields the constraint $p_{sh} + p_{sp} \geq 1$. In this case, the minimal choice is $p_{sp} = 1 - p_{sh}$. 
 Writing $p_{sh}=p$, span-search costs $O(L^{1+p})$ and span-attention attends to $O(L^{1-p})$ tokens per query, 
 for a total cost of $O(L\cdot L^{1-p})=O(L^{2-p})$. Under this tight setting, 
 the overall complexity becomes $\max\{O(L^{1+p}),\, O(L^{2 - p})\}$. 
 To obtain the best asymptotic scaling within this 2-step family, we \emph{balance} the two stages:
\[
1+p = 2-p \quad\Rightarrow\quad p=\tfrac{1}{2}.
\]
At $p=1/2$, both stages scale as $O(L^{3/2})$, and neither dominates; this is the baseline $N=2$ implementation used in this feasibility study.

In practice, one may choose $p_{sh}$ and $p_{sp}$ such that $p_{sh}+p_{sp}>1$ to increase routing redundancy.
In our evaluation, we used $p_{sh} = p_{sp} = 0.54$ to improve empirical performance, trading a higher exponent for a larger routed token budget.
Note that if $p_{sh}=0.54$, then choosing $p_{sp}=0.46$ (the tight setting) yields overall complexity dominated by span-search ($O(L^{1.54})$); increasing $p_{sp}$ to $0.54$ keeps the same overall exponent while increasing the attention-stage budget.
For convenience, one may treat $p_{sh} = p_{sp} = p$ as a simple one-parameter family, with the understanding that the balanced asymptotic optimum for the 2-step mechanism remains $p=1/2$.
This equality choice can be extended to the multi-step case as well, e.g., with $p_{sh} = p_{sp} \ge 1/N$ for $N$ steps.

Finally, in the extremes, $p\to 0$ corresponds to no meaningful routing and recovers dense attention ($O(L^2)$), while $p\to 1$ corresponds to scoring essentially all tokens (also $O(L^2)$).
If we combine the $p\to 1$ case with top-$k$ selection, we recover the standard top-$k$ sparse attention mechanism \cite{liu2025deepseek, xiu2025preliminary}, which also has $O(L^2)$ complexity for exact scoring.
This asymptotic view clarifies where Superlinear attention sits between the dense and top-$k$ extremes, and how choosing $p_{sh}$ and $p_{sp}$ trades compute for routed coverage and redundancy.

\section{Components}
\subsection{Accumulation}
The accumulation component is responsible for efficiently processing 
long sequences and accumulating relevant information up to each token.
An analogous step appears in traditional search algorithms, where data are sorted or preprocessed
to enable efficient search.
One effective way to preprocess is by leveraging linear recurrence mechanisms,
which allow us to compute a summary of the prefix in linear time with respect to the sequence length.
Linear attention \cite{katharopoulos2020transformers} and state space models (SSMs) \cite{gu2021efficiently} are unified under this framework:
both maintain a recurrent state that accumulates key--value associations \cite{dao2024transformers, team2025kimi}.
Formally, we denote the accumulated hidden state $H(t)$ at position $t$ as a function of the prefix $X_{0:t}$:
\begin{equation}
    H(t) = \text{LinearRecurrence}(X_{0:t})
\end{equation}
This hidden state is then projected to produce the key representation $K(t)$, which we reuse as the accumulated key $K_a(t)$ for span-search.
Intuitively, $K_a(t)$ is a cheap, content-dependent summary of the prefix up to $t$, so a high similarity $Q_s(i)\cdot K_a(t)^T$ 
suggests that the candidate span centered at $t$ is promising for query $i$ with search query $Q_s(i)$.
This step ensures that each token has access to a summary of the preceding tokens, which is crucial
for the subsequent span-search step. 

We note that the accumulation step does not need to use linear recurrence specifically. 
We use linear attention/SSM here since it is efficient, widely used, and naturally fits hybrid architectures. 
Other efficient mechanisms can also be used.
The key requirement is that the time complexity 
of the accumulation step should be less than or equal to the overall target complexity of the architecture.
For example, in the baseline $N=2$ implementation, the time complexity of the accumulation step 
should be at most $O(L^{3/2})$. On the other hand, if the overall target complexity is $O(L \log L)$,
then the accumulation step can be at most $O(L \log L)$ as well. The full attention can also be used here, 
but since it is quadratic in complexity, it defeats the purpose of the architecture. 

\subsection{Span search}
The span-search component is the core of the Superlinear attention architecture.
Its primary function is to identify the most relevant spans of the sequence based on the accumulated information from
the previous step. This is achieved by applying a search procedure that efficiently narrows down the search
space to the top-$k$ spans of interest. 

We implement the span-search step by adapting ideas from traditional search algorithms, 
but use accumulated information to guide the search process rather than searching in a sorted array.
Concretely, we use a search query matrix $Q_s$ to 
compute attention scores against the accumulated key matrix $K_a$, 
similar to the attention mechanism in transformers with $S = Q_s K_a^T$.
In our baseline $N=2$ implementation, we reuse the same key matrix used for span-attention as the accumulated key matrix.
However, for span-search we use a dedicated search query matrix $Q_s$.
The attention scores are then used to compare and select the top-$k$ spans of the sequence, 
similar to how traditional search algorithms identify relevant items in a sorted array based on comparison operations.
Specifically, for a query token at position $i$, we compute the search query vector $Q_s(i)$. 
The relevance score $s_{i,t}$ for a candidate anchor span centered at $t$ is calculated as:
\begin{equation}
    s_{i,t} = Q_s(i) \cdot K_a(t)^T
\end{equation}
We then select the set of top-$k$ indices $\mathcal{T}_k(i)$ such that:
\begin{equation}
    \mathcal{T}_k(i) = \textsc{TopK}\big(\textsc{Anchors}(i),\, s_{i,t},\, k\big)
\end{equation}

In our baseline $N=2$ instantiation, $\textsc{Anchors}(i)$ is a fixed stride pattern that yields $|\textsc{Anchors}(i)|=O(i^{1/2})$ candidates per query. More generally, one can choose an anchor schedule with $|\textsc{Anchors}(i)|=O(i^p)$ for a tunable exponent $p\in(0,1)$. Previously, we implemented an $O(L^{1+p})$ attention mechanism that attends to $O(L^p)$ tokens, 
which can be adapted for the span-search step. For the baseline $N=2$ implementation,
we set $p=1/2$, yielding $O(L^{1/2})$ anchors per query and $O(L^{3/2})$ total span-search complexity.

For multi-step span-search with $N$ steps, the target is that the final routed span length scales as $O(L^{1/N})$.
We can describe the mechanism in terms of a recursively shrinking candidate region.
Let $R_0$ be the full prefix (length $|R_0|=O(L)$), and let $R_m$ denote the candidate region after $m$ span-search steps.
We design each step to reduce region length by a factor of $L^{1/N}$:
\[
|R_m| = O\big(L^{(N-m)/N}\big) \quad\text{for } m=0,1,\dots,N-1.
\]
At step $m$ (operating on a region of length $|R_{m-1}|$), we score a sublinear set of anchors of size
$O\big(|R_{m-1}|^{1/(N-m+1)}\big)=O(L^{1/N})$ and select a subregion of length
$|R_m|=O\big(|R_{m-1}|^{(N-m)/(N-m+1)}\big)=O\big(L^{(N-m)/N}\big)$.
Therefore, each span-search step scores only $O(L^{1/N})$ anchors per query, and the overall span-search work across $L$ queries remains $O(L\cdot L^{1/N})=O(L^{1+1/N})$ up to constant factors in $N$.

Alternatively, one may use a binary or $k$-ary search strategy with $O(\log L)$ steps, which yields $O(L \log L)$ complexity.
In this approach, each step reduces the search space by half or by a factor of $k$, respectively.
This method allows us to efficiently narrow down the relevant spans of the sequence while maintaining random context access. 
However, the implementation of such binary or $k$-ary search is more complex and is left for future work. 

\begin{figure}[t]
    \centering
    \makebox[\textwidth][c]{%
    \includegraphics[width=1.3\textwidth]{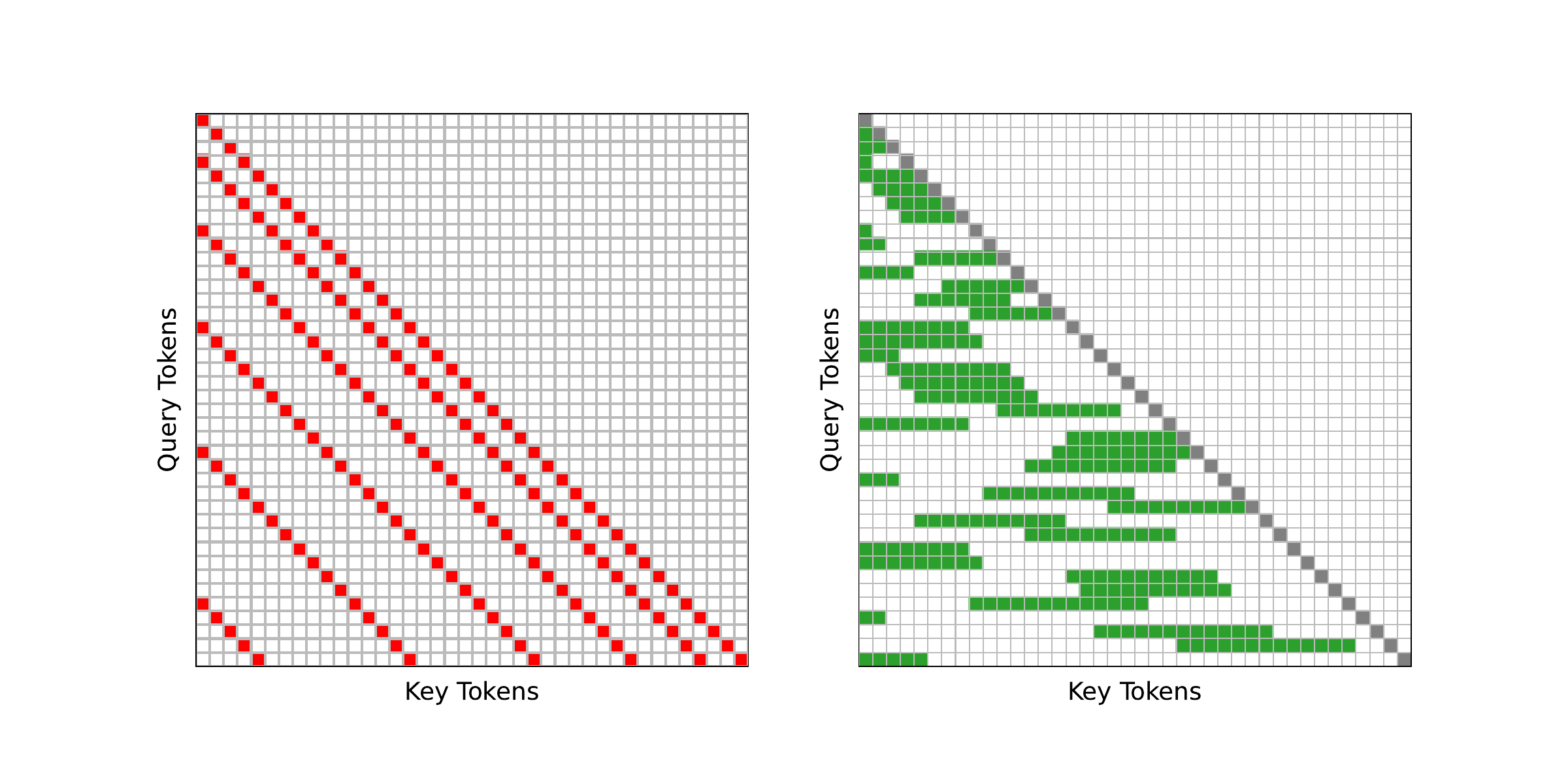}
    }
    \caption{Left: Stride pattern for the span-search step with $N=2$, where the search complexity is $O(L^{3/2})$. 
    Right: Illustration of the selected spans for the span-attention step, which also has a complexity of $O(L^{3/2})$. 
    The span tokens are shown in green, while the diagonal tokens are shown in gray. 
    Each row corresponds to the attention map for a single query token. Figure~\ref{fig:superlinear_overview} 
    shows how each row fits into the Superlinear attention architecture.
    In actual implementations, 
    the diagonal tokens are expanded into sliding windows to capture local dependencies, and 
    multiple spans are used and combined to improve expressiveness and make the span-search trainable via backpropagation.
    Combined, the span-search and span-attention steps achieve an overall complexity 
    of $O(L^{3/2})$ while ensuring random context access.}
    \label{fig:span_search_and_attention}
\end{figure}

\subsection{Span attention}

\begin{figure}[t]
\centering
    \makebox[\textwidth][c]{%
    \includegraphics[width=0.8\textwidth]{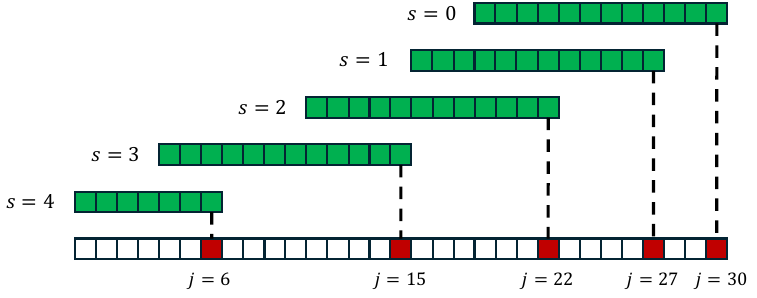}
    }
\caption{
Illustration of random context access (structural non-exclusion) induced by the $N=2$ span construction. 
In this example, query position is $i=30$, the backward extent is $b=2$ and the forward extent is $f=0$, the span length is therefore 
$(b+f) \cdot \lceil i^{1/2} \rceil = 2 \cdot \lceil\sqrt{30}\rceil = 12$. 
$j$ is the key position index, and $s$ is the stride index for the span-search step.
Then the anchor tokens for search are located at $j=i - (s+1)^2 + 1$, or $30, 27, 22, 15, 6$ for $s=0, 1, 2, 3, 4$ respectively.
As shown, all token positions from $0$ to $30$ fall inside at least one \emph{candidate} span (centered at some anchor), meaning each position is reachable if the routing selects an appropriate anchor.
}
\label{fig:structural_non_exclusion}
\end{figure}

Once the top-$k$ spans have been identified through the span-search step,
the span-attention component performs standard attention on these spans.
For each selected anchor $t \in \mathcal{T}_k(i)$, we define a span of tokens $S_t$ centered at $t$. 
We then perform standard scaled dot-product attention within this span to obtain the span-specific output $A_{i,t}$:
\begin{equation}
    A_{i,t} = \text{Attention}(Q(i), \{K_j\}_{j \in S_t}, \{V_j\}_{j \in S_t})
\end{equation}
This allows the model to focus on the most relevant parts of the sequence,
enabling it to capture important dependencies and relationships between tokens.

While the span-attention step processes $O(L^{1/N})$ tokens per query, 
the span length must be chosen to ensure random context access.
We derive the span length for the baseline $N=2$ implementation below; the same argument extends to multi-step span-attention.

For $N=2$, using the stride pattern shown in Figure~\ref{fig:span_search_and_attention}, 
for general $p$, query position $i$, and stride index $s=0, 1, 2, ...$, the anchor tokens for search are located at:
$i-(s+1)^{1/p}+1$. To verify that this corresponds to $O(i^p)$ tokens, we can calculate the maximum stride index $S$ such that:
$ (S+1)^{1/p} \leq i $ which gives $S = O(i^p)$. 

Using these anchor locations for search, for $p=1/2$ the anchor tokens are located at $i, i-3, i-8, i-15, ...$. In this setting, the gap 
between two adjacent anchor tokens is: $3, 5, 7, ...$. In general, the gap between two adjacent anchor tokens is:
$(s+1)^{1/p} - s^{1/p}$. To obtain the lower bound on the span length, we can 
consider the mean value theorem for the function $g(x)=x^{1/p}$. Then there exists $\xi\in(s,s+1)$ such that
\[
(s+1)^{1/p} - s^{1/p} = g'(\xi) = \tfrac{1}{p}\,\xi^{\frac{1}{p}-1} \le \tfrac{1}{p}(s+1)^{\frac{1}{p}-1}.
\]
Since the largest stride index is $S=O(i^p)$, the maximum anchor gap satisfies
\[
\Delta_{\max} \le \tfrac{1}{p}(S+1)^{\frac{1}{p}-1} = O\big(i^{1-p}\big).
\]
Therefore, if we choose a base span length $\ell(i)=\lceil i^{1-p}\rceil$ and 
set the backward span length to $b\,\ell(i)$ with a constant $b\ge \tfrac{1}{p}$,
then spans centered at anchors are wide enough to cover the largest anchor gaps. 
Equivalently, for every eligible key $j\le i$, 
there exists at least one anchor $t\in\textsc{Anchors}(i)$ such that $j\in S_t$, so $j$ is reachable by routing to that anchor.
In particular, for $p=1/2$ we obtain the natural condition $b\ge 2$, matching the example in Figure~\ref{fig:structural_non_exclusion}.

The coefficient $b$ determines the backward extent of the span from the anchor token. To increase expressiveness,
we also introduce a forward extent $f$ to the span, which extends the span forward from the anchor token.
This allows the model to attend to tokens that are located after the anchor token, increasing the range of tokens that can be attended to.
This is to ensure that meaningful spans are captured properly without being cut off by the anchor token at 
arbitrary locations enforced by the stride pattern. One close analogy is the extra overlap added in RAG 
retrieval to ensure that meaningful chunks are not cut off arbitrarily.

Although one can set a dynamic span length based on the stride index $s$, 
where the maximum gap scales as $\Delta_{\max}=O(i^{1-p})$,
for simplicity while maintaining the overall complexity, we use a fixed span length of $O(i^{1-p})$ for all spans at query position $i$.
Therefore, the span length for position $i$ is $(b+f) \cdot \lceil i^{1-p} \rceil$.

A potential concern is that attending to only a subset of tokens via the top-$k$ spans loses information.
However, the overall architecture is a modification of the multi-headed self-attention mechanism. 
Each head can attend to $k$ different spans of the sequence, and by combining multiple heads, 
the model can still capture a wide range of dependencies across the entire sequence.
Furthermore, since the span-search step is trainable via backpropagation, 
the model can learn to identify the most relevant spans for attention,
ensuring that important information is not overlooked.

In addition, sliding window attention is combined with span-attention
to capture local dependencies. Since sliding window attention is linear in complexity,
it does not affect the overall complexity of the architecture. The sliding window is chosen 
based on the stride index, so that span-search happens immediately after a sliding window attention block 
to avoid gaps between the spans and the sliding windows.

\subsection{Combination}
\label{sec:combination}
The combination component integrates the attended spans from the span-attention step
using a differentiable gating mechanism similar to Mixture-of-Experts (MoE) routing \cite{shazeer2017outrageously, fedus2022switch}.

The final output $O_i$ for query $i$ is a weighted sum of the span-attention outputs, 
weighted by the normalized search scores:
\begin{equation}
    O_i = \sum_{t \in \mathcal{T}_k(i)} \alpha_{i,t} A_{i,t}
\end{equation}
where the gating weights $\alpha_{i,t}$ are obtained by applying softmax to the search scores:
\begin{equation}
    \alpha_{i,t} = \frac{\exp(s_{i,t})}{\sum_{t' \in \mathcal{T}_k(i)} \exp(s_{i,t'})}
\end{equation}
Here, $s_{i,t}$ corresponds to the search score for span $t$, 
and $A_{i,t}$ is the attended output from the span-attention step for span $t$.

The softmax normalization ensures that the most relevant spans have greater influence on the final output.
More importantly, the softmax-weighted combination is differentiable,
enabling the span-search scoring to be trained via backpropagation---analogous to how routing decisions in MoE models
are learned through gradient descent.

In our implementation, the discrete top-$k$ selection is treated as non-differentiable, and gradients flow through the softmax weights $\alpha_{i,t}$ over the selected spans. This means unselected spans receive zero gradient on that update, so credit assignment occurs only through spans that are selected.
Routing redundancy (e.g., larger $k$ and larger span extents) can increase the probability that useful spans are selected early in training, which may ease optimization by helping gradient descent discover and reinforce effective routing behaviors.

The computed scores must correspond to the probability of selecting the spans. 
In the case of $N=2$ implementation, the score computation is straightforward since there is only one span-search step.
However, in the case of multi-step span-search with $N>2$, the score computation becomes more complex.
In this case, the final score for each span must take into account the scores from all previous span-search steps.
One simple approach is to view each step as producing a (log-)score over subregions and compose scores across steps (e.g., by summing log-scores along the selected path). The exact factorization depends on how intermediate regions are defined; we leave a detailed treatment to future work.

\subsection{Pseudocode}

Algorithm~\ref{alg:superlinear_n2} summarizes the $N=2$ case. The first step is a 
content-dependent \emph{span-search}: for each query position $i$, 
we score a sublinear set of candidate anchors $t\in\textsc{Anchors}(i)$ using a 
routing vector $Q_s(i)$ and a per-position representative stream $K_a(t)$. 
We then select the top-$k$ anchors and attend to contiguous spans centered at those anchors.

The representative stream $K_a$ can be implemented in several ways. 
Conceptually, it summarizes the prefix up to each position so that scoring anchors is cheap; 
it can be produced by an accumulation module 
(e.g., a linear attention or SSM layer) and cached. 
In our baseline $N=2$ implementation, the hidden state from the preceding Mamba-2 layer is projected into the key matrix $K(t)$,
and we reuse $K_a(t) = K(t)$ for span-search scoring, which keeps the KV-cache interface unchanged.

For each selected anchor $t$, we form a routed span $S_t$ of length $(b+f)\,\ell(i)$ around $t$, with backward length $b\,\ell(i)$ and forward length $f\,\ell(i)$
(e.g., $\ell(i)\propto i^{1/2}$ for the $N=2$ / $O(L^{3/2})$ setting; in our implementation, $b$ and $f$ correspond to \texttt{backward\_factor} and \texttt{forward\_factor}). 
Optionally, we also include a local causal window $W$ to preserve short-range modeling. 
When $W$ is used, we exclude window indices from anchor candidates and merge $S_t$ with $W$ (de-duplicated) to avoid redundant work.

\begin{algorithm}[t]
\caption{Superlinear attention (N=2)}
\label{alg:superlinear_n2}
\begin{algorithmic}[1]
\Require Tokens $X[0..L\! -\! 1]$; causal mask ($j\le i$); window size $w$ (optional)
\Require Candidate anchors $\textsc{Anchors}(i)$ with $|\textsc{Anchors}(i)|=O(i^{1/2})$; top-$k$; span extents $(b,f)$; base span length $\ell(i)$ (e.g., $\ell(i)=\lceil i^{1/2}\rceil$)
\State Precompute $Q(i),K(i),V(i)$; routing vectors $Q_s(i)$; representatives $K_a(t)$
\For{$i=0$ to $L\! -\! 1$}
    \State $W \gets [\max(0, i-w+1),\, i]$ \Comment{treat as empty if $w=0$}
    \State $T \gets \textsc{Anchors}(i) \setminus W$
    \ForAll{$t\in T$}
        \State $s_t \gets \langle Q_s(i), K_a(t)\rangle$
    \EndFor
    \State $\mathcal{T}_k \gets \textsc{TopK}(T, s_t, k)$
    \If{$\mathcal{T}_k$ is empty}
        \State $O[i] \gets \textsc{Attention}(Q(i), K[W], V[W])$; \textbf{continue}
    \EndIf
    \ForAll{$t\in \mathcal{T}_k$}
        \State $S_t \gets [\max(0, t-b\,\ell(i)),\, \min(i, t+f\,\ell(i))]$
        \State $J_t \gets \textsc{DedupConcat}(S_t, W)$
        \State $A_t \gets \textsc{Attention}(Q(i), K[J_t], V[J_t])$
    \EndFor
    \State $\alpha_t \gets \textsc{Softmax}\big(\{s_t\}_{t\in\mathcal{T}_k}\big)$
    \State $O[i] \gets \sum_{t\in\mathcal{T}_k} \alpha_t\, A_t$
\EndFor
\State \Return $O$
\end{algorithmic}
\end{algorithm}

\section{Experiments (Feasibility Study)}

\subsection{Evaluation scope and limitations}
Our experiments are a feasibility study: they are intended as initial validation of architectural and systems viability rather than a comprehensive quality study.
We emphasize (i) scaling behavior and throughput at very long context lengths, where dense attention becomes impractical, and (ii) a focused learnability check on a long-context retrieval task (NIAH \cite{kamradt2023needle}).
We do not present extensive ablations over routing schedules, hyperparameters, model sizes, or datasets, and we do not claim state-of-the-art accuracy.
Instead, we aim to establish that the proposed mechanism is (a) structurally random-context-access-preserving, (b) asymptotically subquadratic in the attention components, and (c) implementable with practical performance at very long context lengths.

\subsection{Baseline N=2 implementation}
To validate the effectiveness of the proposed Superlinear attention architecture,
we implement a baseline $N=2$ variant for feasibility evaluation. This implementation achieves $O(L^{3/2})$ complexity while maintaining random context access.

Since our architecture combines an accumulation step with a span-search + span-attention mechanism, we implement it by modifying an existing hybrid transformer.
Specifically, we replace all standard attention layers with Superlinear attention layers while keeping the rest of the model unchanged.
This isolates the effect of the proposed mechanism while preserving the overall model design.

For our baseline $N=2$ implementation and experimentation, we utilize one of the latest hybrid transformer models, 
NVIDIA-Nemotron-3-Nano-30B-A3B \cite{nvidia2025nemotron3nanoopen,nvidia2025nvidianemotron3efficient}, 
as our base architecture. The Nemotron-3-Nano model combines hybrid attention mechanisms with MoE layers to 
achieve efficient processing of long sequences. By integrating our Superlinear attention layer into this architecture,
we can effectively evaluate its performance and efficiency in a real-world setting. 
Figure~\ref{fig:superlinear_overview} shows the overall architecture with 
the Superlinear attention layer integrated into the Nemotron-3-Nano model.

Our Superlinear attention mechanism is compatible with a variety of positional encoding choices, but in this work we follow the base model configuration.
Nemotron-3-Nano uses NoPE (no position embedding), and we keep this setting unchanged.
Sequence order still enters through the standard causal eligibility constraint ($j\le i$) 
and through the order-sensitive accumulation module (Mamba-2 / linear recurrence) 
that produces the representative stream used for routing and attention.

In the Nemotron-3-Nano architecture, each standard attention layer is preceded by a Mamba-2 layer \cite{dao2024transformers}.
Mamba-2, like other SSMs and linear attention variants, maintains a recurrent state that accumulates information from the prefix.
This naturally serves as our accumulation step: the Mamba-2 hidden state is projected into the key matrix $K$,
which we then reuse as $K_a$ for span-search scoring.
For the standard attention layer that we are replacing, we introduce a dedicated search query matrix $Q_s$ for the span-search step.
The other three components (span-attention, combination, and output projection) are implemented as a single attention block. 
It replaces the standard attention layer with the same input and output interface, making it easy to integrate into the existing architecture.
Since $K_a = K$, the KV-cache interface remains unchanged.

In our comparative studies, we keep the non-attention components (e.g., the accumulation/Mamba and MoE layers) 
fixed across baselines and focus on isolating the effect of the attention kernel 
and its scaling behavior with context length.

\subsection{Kernel design}
While the Superlinear attention architecture reduces the theoretical complexity to $O(L^{3/2})$, 
implementing it efficiently on GPUs presents unique challenges due to the irregular sparsity pattern. 
Unlike block-sparse or sliding window attention, the span-search mechanism produces content-dependent spans
where each query attends to a different, potentially dispersed set of key tokens.
In practice, the systems challenge is concentrated in prefilling and training, where all $L$ queries must be processed and span footprints vary across queries; by contrast, incremental decoding evaluates a single query per step and can be handled with a comparatively simple path built from standard primitives.

\begin{figure}[H]
    \centering
    \makebox[\textwidth][c]{%
    \includegraphics[width=1.0\textwidth]{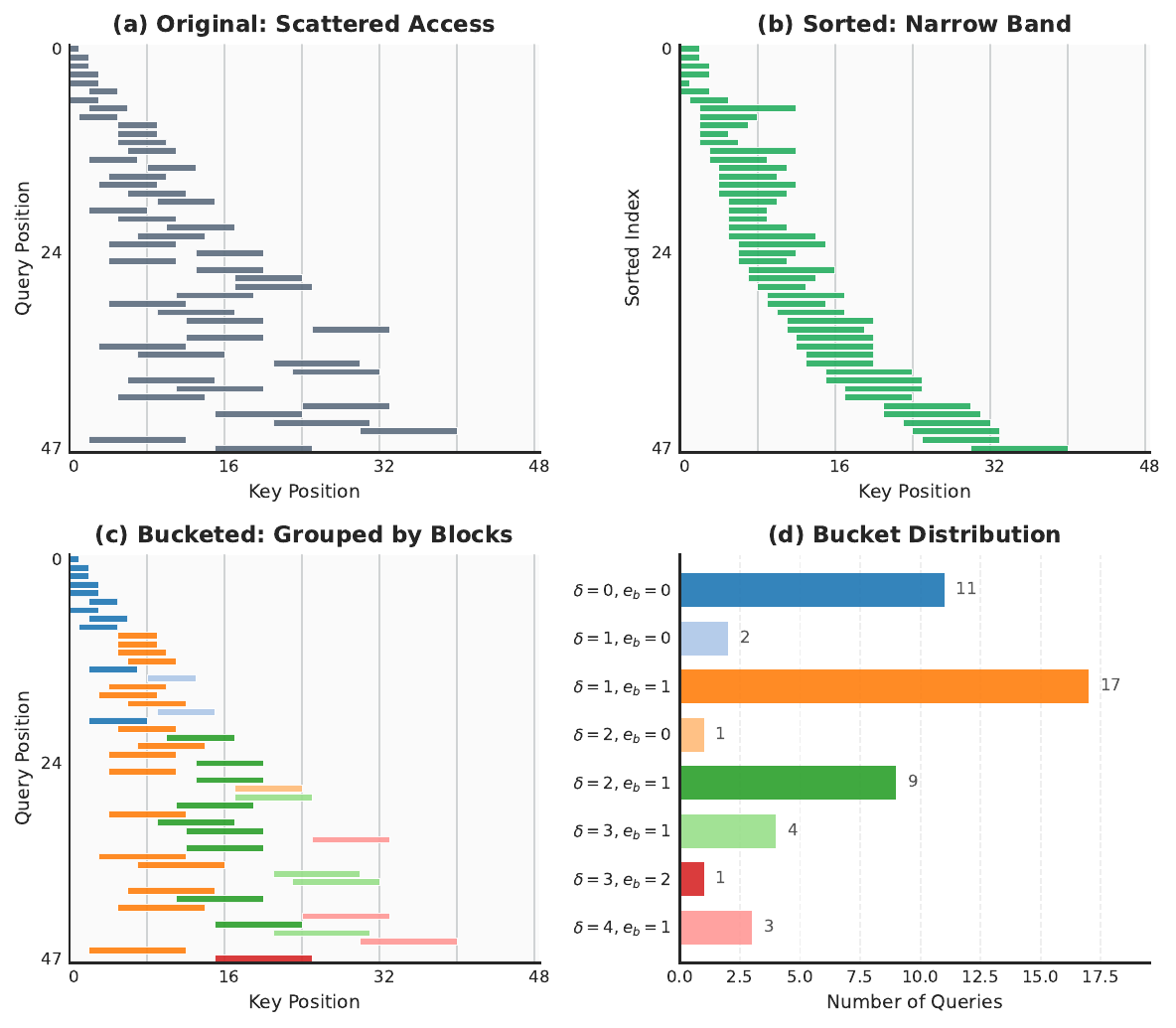}
    }
    \caption{Comparison of kernel designs for irregular span-attention patterns. 
    (a) The original scattered span-attention pattern.
    (b) The sorting approach reorders queries to create a contiguous diagonal band, 
    but incurs $O(L \log L)$ overhead from sorting and permutation. 
    (c) The proposed bucketed approach groups queries by their key-block footprint (color-coded) into buckets.
    (d) Distribution of queries per bucket (key-block range). The kernel uses dynamic work-stealing to handle varying bucket sizes, ensuring high GPU occupancy.
    Queries in the same bucket are processed together, ensuring efficient memory access without the need for global sorting. This design is critical for realizing the theoretical speedup of the Superlinear attention mechanism.}
    \label{fig:bucketed_kernel_manuscript}
\end{figure}

\paragraph{Prefilling/training kernel.}
During prefilling and training, all $L$ queries must be processed, and the main challenge is the irregular sparsity pattern across queries.
Figure~\ref{fig:bucketed_kernel_manuscript}(a) illustrates this scattered access pattern.

One approach to improve locality is to sort queries by their span start positions. 
This transformation groups similar spans together, creating a narrow diagonal structure (Figure~\ref{fig:bucketed_kernel_manuscript}(b)), 
which improves memory locality. However, global sorting introduces an $O(L \log L)$ overhead and requires permuting query and output tensors, which can dominate at moderate sequence lengths.

To address this, we developed a \textbf{bucketed kernel design} that effectively handles irregular spans without global sorting. 
The core idea is to group (query, span) pairs into ``buckets'' based on their key-block footprint 
(Figure~\ref{fig:bucketed_kernel_manuscript}(c)).
Specifically, we divide the key sequence into fixed-size blocks (e.g., 64 tokens)
and assign each query to a bucket defined by its span's end-block index and length. 
Queries within the same bucket share the same memory access pattern for keys and values, 
allowing them to be processed together efficiently in a single kernel tile. 
This approach maximizes GPU utilization through work-stealing and atomic counters, 
avoiding the need for expensive sorting or padding while accommodating the dynamic nature of span-attention, as shown in the distribution of bucket sizes (Figure~\ref{fig:bucketed_kernel_manuscript}(d)).

\paragraph{Decoding path.}
Incremental decoding is considerably simpler because each step evaluates a single query position, so the cross-query irregularity that motivates bucketing in prefilling is absent. In this regime, we compute routing scores over the anchor set for the current position and evaluate attention over the resulting routed spans. In our implementation, this path is expressed using standard tensor operations and existing fused attention primitives (e.g., SDPA) rather than a bespoke decode-specific kernel, which keeps the implementation lightweight while retaining the $O(L^{1/2})$ per-step attention scaling implied by the $N=2$ design.

\subsection{Inference and computational efficiency}
We benchmark the baseline $N=2$ implementation against a dense-attention baseline to assess 
whether the theoretical $O(L^{3/2})$ advantage translates into practical speedups in our implementation.
We compare our Superlinear attention kernel (Figure~\ref{fig:bucketed_kernel_manuscript}) to FlashAttention-2 under the same model configuration.
For prefilling, we report throughput in tokens/sec.
For decoding, we report decode throughput in tokens/sec as a function of context length.

\begin{figure}[t]
    \centering
    \makebox[\textwidth][c]{%
    \includegraphics[width=1.0\textwidth]{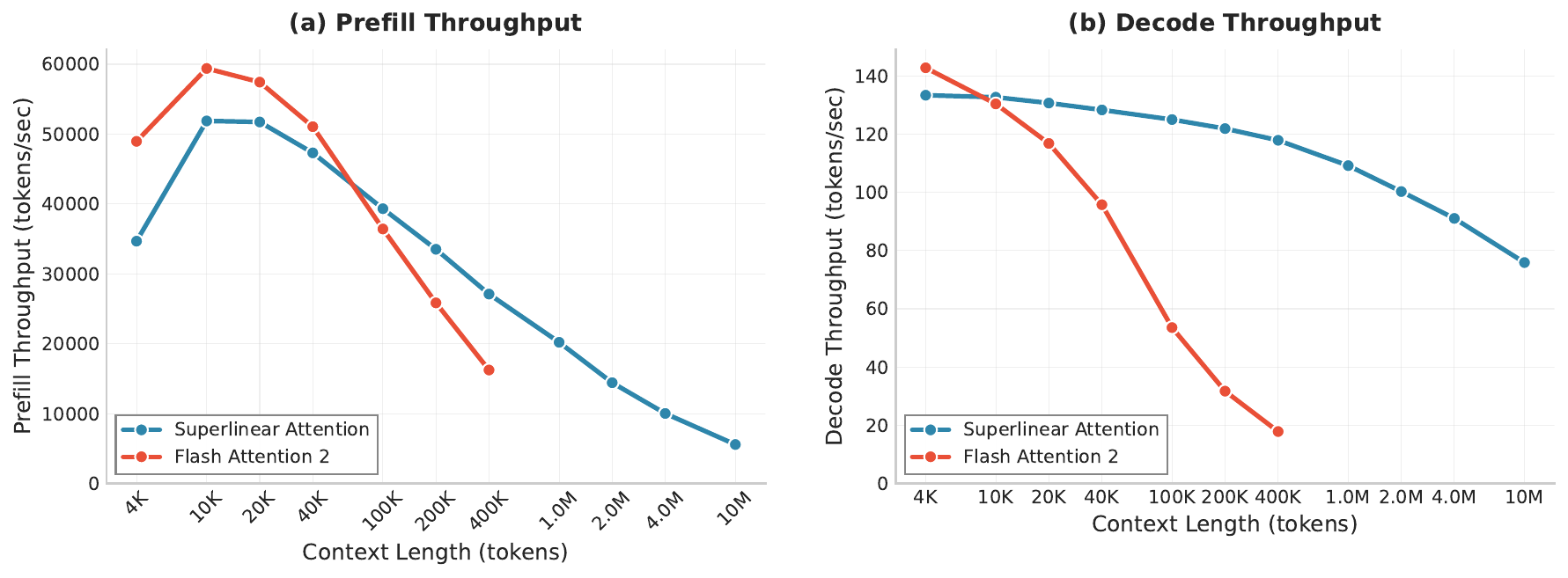}
    }
    \caption{Throughput comparison between Superlinear attention (our baseline $N=2$ implementation) and FlashAttention-2.
    (a) Prefill throughput (tokens/sec) under chunked prefilling (32K chunk size) versus context length.
    (b) Decode throughput (tokens/sec) versus context length.
    Superlinear attention scales more gracefully with context and enables efficient multi-million-token inference.}
    \label{fig:throughput_comparison_flash_vs_superlinear}
\end{figure}

Figure~\ref{fig:throughput_comparison_flash_vs_superlinear} summarizes the results.
In prefilling (Figure~\ref{fig:throughput_comparison_flash_vs_superlinear}(a)), 
FlashAttention-2 is slightly faster at short contexts, but its throughput decreases rapidly with context length.
Superlinear attention scales more gracefully, overtaking FlashAttention-2 near 60K tokens.
At 100K--400K, Superlinear attention achieves substantially higher throughput,
 and it continues to run efficiently into the multi-million-token regime (1M--10M), 
 where dense attention becomes impractical. At 10M tokens, we measure a prefill throughput of 5576 tokens/sec (roughly 30 minutes to prefill 10M tokens).
This demonstrates that the proposed kernel remains usable in regimes where exact dense 
attention would be computationally prohibitive. At these lengths, 
end-to-end throughput reflects a combination of attention, memory bandwidth, and other model components, 
but crucially the attention cost no longer scales quadratically with context.

In decoding (Figure~\ref{fig:throughput_comparison_flash_vs_superlinear}(b)), the difference is more pronounced.
FlashAttention-2 decode throughput drops sharply as context grows (reflecting the increasing cost of attending over a long prefix),
whereas Superlinear attention maintains relatively stable decode throughput across a wide range of context lengths.
Even at multi-million-token contexts, decode performance reflects a combination of routed attention work 
(KV-cache reads over routed spans, scaling as $O(L^{1/2})$ for $N=2$), $O(1)$ per-token model components 
(e.g., cached recurrence/MLP/MoE compute), and memory bandwidth effects.
At 10M tokens, decode latency is 13.18 ms/token (equivalently about 76 tokens/sec).
This level of per-token latency keeps generation interactive even at ten-million-token contexts.

Practically, we can employ a hybrid strategy in production: at short contexts, 
we fall back to dense attention (e.g., SDPA/FlashAttention) to avoid kernel overheads;
at longer contexts, we switch to the Superlinear attention kernel to realize the asymptotic gains.

\paragraph{Experimental setup.}
All throughput results in this section are measured on a single NVIDIA B200 GPU (180GB VRAM) with batch size 1 (no packed batches), reflecting a common use case for extremely long-context inference.
We report end-to-end throughput (including non-attention components such as accumulation/Mamba and MoE).
For prefilling, we use chunked prefilling with a chunk size of 32K tokens.
For decoding, we report average tokens/sec computed over generating 20 tokens (averaged across these 20 decode steps) after a short warmup.

\paragraph{Kernel configuration.}
For the Superlinear attention results reported here, we use a key-block size of 64 tokens for bucketing,
sliding window size \texttt{w}=1088 for local attention,
 \texttt{topK} = 2, and redundancy factors \texttt{backward\_factor} = 4 and \texttt{forward\_factor} = 2.
Unless otherwise stated, we set \texttt{search\_exponent} = \texttt{span\_exponent} = 0.5, 
matching the \emph{Extended} configuration used in the training results below.

\FloatBarrier

\subsection{Training and model performance}

Because the original Nemotron-3-Nano model does not include the search query matrix $Q_s$ for the span-search step,
we initialize $Q_s$ using the same scheme as the other query matrices in the model.
This ensures that $Q_s$ starts with values that are compatible with the rest of the architecture.

Even with this initialization, fine-tuning is required to adapt the model to the Superlinear attention mechanism.
The current study targets architectural validation rather than state-of-the-art performance.
A key question is whether the model can effectively learn to utilize the Superlinear attention mechanism
for long-context tasks. For this reason, we focus on directly fine-tuning the modified Nemotron-3-Nano model
on the needle-in-a-haystack (NIAH) task from the RULER benchmark \cite{hsieh2024ruler}, which builds on the original NIAH pressure test popularized by Kamradt \cite{kamradt2023needle}, to show that it is possible to train the model for long-context tasks.

Although the proposed architecture is trainable via backpropagation, the model must still identify a useful span
before it can attend to it effectively. This is non-trivial, especially when the model is not pre-trained with the Superlinear attention mechanism.
To address this, we use curriculum learning \cite{bengio2009curriculum} by starting with shorter contexts and gradually increasing the context length during fine-tuning.
We start with 4K context and increase to 64K over the course of fine-tuning.

A further challenge is that the model must learn to accumulate relevant information effectively in the accumulation step for span-search to work well.
While the model works well on the context lengths it was trained on, it struggles to generalize to longer contexts. A solution 
to this problem is to use different exponents for the search and attention steps during fine-tuning. Instead of using $p=1/2$ for both steps,
we use a smaller exponent for the search step (e.g., $p=1/3$) and a larger exponent for the attention step (e.g., $p=2/3$).
This has two effects: (1) the search step has lower complexity, which can make it easier for the model to learn to identify relevant spans,
and (2) the attention step has higher complexity, which allows the model to attend to more tokens within the selected spans.
Together, these changes can improve generalization to longer contexts by making accumulation and routing easier to learn.

In our implementation, these exponent knobs correspond to \texttt{search\_exponent} (span-search) and \texttt{span\_exponent} (span-attention).
In addition, we use simple multiplicative ``factor'' knobs to increase routing redundancy (e.g., scoring more candidates and/or keeping more spans per query).
These factors increase compute and memory proportionally, but can stabilize training by making it easier to discover and reinforce useful spans early in fine-tuning.

\begin{figure}[t]
    \centering
    \makebox[\textwidth][c]{%
    \includegraphics[width=1.0\textwidth]{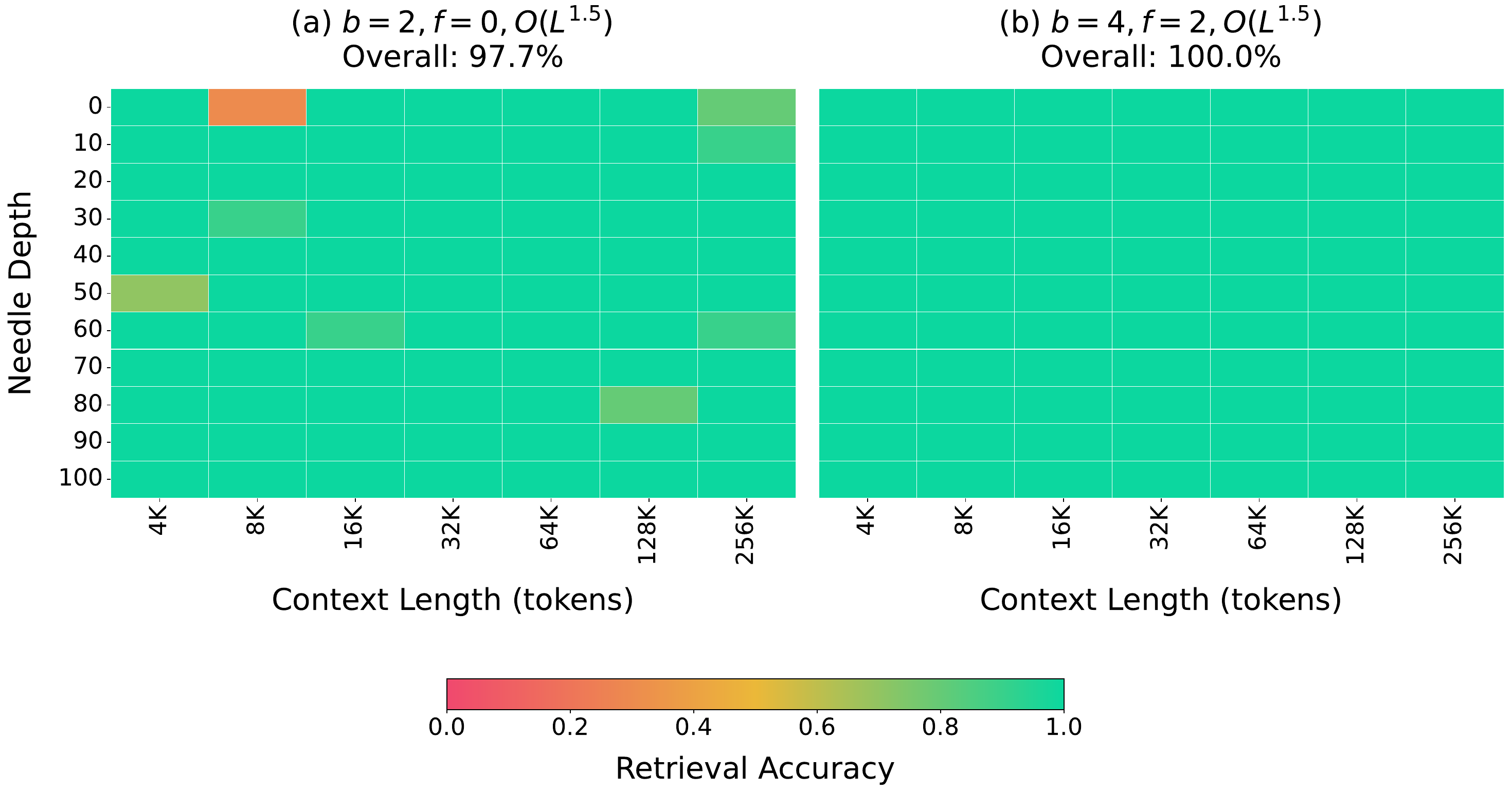}
    }
    \caption{NIAH retrieval accuracy heatmaps.
    Columns show context lengths (4K--256K tokens) and rows show needle depth (0--100\% into the context).
    (a) Baseline.
    (b) Extended (+2/+2 factors).}
    \label{fig:niah_comparison_heatmaps}
\end{figure}

To quantify how these training choices affect long-context retrieval, we evaluate on the RULER needle-in-a-haystack (NIAH) benchmark.
Figure~\ref{fig:niah_comparison_heatmaps} reports retrieval accuracy across a grid of context lengths and needle depths.
We compare a baseline setting (``Baseline'') and an extended routing configuration (``Extended'', +2/+2 factors).
Concretely, \emph{Baseline} uses \texttt{search\_exponent} = \texttt{span\_exponent} = 0.50 with \texttt{backward\_factor} = 2.0 
and \texttt{forward\_factor} = 0; \emph{Extended} keeps exponents fixed and increases 
routing redundancy to \texttt{backward\_factor} = 4.0 and \texttt{forward\_factor} = 0.2 (i.e., ``+2/+2 factors'' relative to Baseline).
Becaus the backward factor must be at least 2.0 to ensure random context access for $p=1/2$,
we extend the forward factor by the same amount (2.0) as the minimal backward factor so that it covers a similar range of tokens beyond the anchor.
In addition, the backward factor is also extended by 2.0 to increase redundancy in the backward direction.
Both configurations use \texttt{topK} = 2 and a key-block size of 64 tokens for bucketing; only exponents and redundancy factors are changed.
In the baseline setting, performance is uneven across the grid, with clear failure cases at shallow and mid depths for longer contexts (e.g., 256K) and even some shorter contexts (e.g., 8K).
Adding redundancy (Extended) reduces the frequency of these failures but does not fully remove the worst-case drops.

\FloatBarrier

In this work, our goal is architectural validation rather than peak benchmark performance.
Accordingly, we perform end-to-end fine-tuning (no frozen components) so that the accumulation, routing, and span-attention behaviors can co-adapt.
An important direction for future work is to provide more direct learning signal to the span-search/router (e.g., auxiliary losses or task-specific supervision) and to evaluate on a broader suite of long-context tasks beyond NIAH.

\section{Conclusion}
We introduced \textbf{Superlinear attention}, a fully trainable multi-step attention formulation that reduces the asymptotic cost of long-context processing while preserving \emph{random context access}: any eligible token position can be selected by the content-dependent routing mechanism.
In the minimal $N=2$ instantiation, span-search and span-attention each scale as $O(L^{3/2})$, and the architecture can be integrated into an existing hybrid transformer with minimal interface changes.

To make this practical on modern GPUs, we described a bucket-based kernel design that groups irregular spans by key-block footprint, enabling efficient tiling and high occupancy without global sorting.
Empirically, our Superlinear attention implementation shows substantial prefill and decode speedups at long contexts in our testbed, and remains usable in regimes where dense attention becomes impractical.

On the training side, we showed that the routing and accumulation components can be learned end-to-end via fine-tuning.
In particular, curriculum learning \cite{bengio2009curriculum} and small adjustments to the span-search/span-attention exponents improve extrapolation to longer contexts, and the resulting models achieve strong long-context retrieval accuracy on NIAH across a wide grid of lengths and depths.

This work leaves several important directions open.
First, broader evaluation on diverse long-context benchmarks is needed to characterize quality/efficiency tradeoffs beyond retrieval.
Second, more direct supervision for routing (e.g., auxiliary objectives) may further stabilize training and reduce sample complexity.
Finally, extending the approach to more than two steps and co-designing kernels for those settings could push the complexity closer to $O(L^{1+1/N})$ for larger $N$, while retaining the random context access properties that motivate the architecture.

\vspace{0.5em}
\noindent\textbf{Positioning.} This paper is intended as an architecture-and-systems feasibility report: it introduces the mechanism, formalizes random context access as structural non-exclusion, and demonstrates that the resulting irregular span pattern can be implemented efficiently at multi-million-token contexts.
A companion empirical study will focus on broader task-level quality evaluation and ablations across routing and training choices.

\section*{Acknowledgments}
The authors have filed patent applications related to aspects of the methods described in this work.

\bibliographystyle{plainnat}
\bibliography{references}

\end{document}